\documentclass{article}

\usepackage{arxiv}
\renewcommand{\date}[1]{} 
\usepackage[utf8]{inputenc}   
\usepackage[T1]{fontenc}      
\usepackage{CJKutf8}          

\usepackage[utf8]{inputenc} 
\usepackage[T1]{fontenc}    
\usepackage{hyperref}       
\usepackage{url}            
\usepackage{booktabs}       
\usepackage{amsfonts}       
\usepackage{nicefrac}       
\usepackage{microtype}      
\usepackage{lipsum}
\usepackage{graphicx}
\graphicspath{ {./images/} }
\usepackage{booktabs} 
\usepackage{subcaption}
\usepackage{appendix}

\title{Automatic database description generation for Text-to-SQL}

\author{Yingqi Gao, \textbf{Zhiling Luo}\thanks{Corresponding Author, godot.lzl@alibaba-inc.com}
\\
\\ Alibaba Group
}

\begin{document}
\maketitle

\begin{abstract}

In the context of the Text-to-SQL task, table and column descriptions are crucial for bridging the gap between natural language and database schema. 
This report proposes a method for automatically generating effective database descriptions when explicit descriptions are unavailable. 
The proposed method employs a dual-process approach: a coarse-to-fine process, followed by a fine-to-coarse process.
The coarse-to-fine approach leverages the inherent knowledge of LLM to guide the understanding process from databases to tables and finally to columns. 
This approach provides a holistic understanding of the database structure and ensures contextual alignment. 
Conversely, the fine-to-coarse approach starts at the column level, offering a more accurate and nuanced understanding when stepping back to the table level.
Experimental results on the Bird benchmark indicate that using descriptions generated by the proposed improves SQL generation accuracy by 0.93\% compared to not using descriptions, and achieves 37\% of human-level performance.
The source code is publicly available at~\url{https://github.com/XGenerationLab/XiYan-DBDescGen}.
\end{abstract}

\section{Introduction}
The technology that converts natural language queries into structured query language (SQL), known as Text-to-SQL or NL2SQL, allows both non-experts and advanced users to easily extract insights from complex datasets\cite{chen2023text, liu2024survey, tai2023exploring}.

In the context of the NL2SQL task, table and column descriptions provide semantic context that helps the model understand the relationships between various entities in the user's query and the database schema.
This context is crucial for accurately generating SQL queries.
Table descriptions provide a high-level understanding of what each table represents, guiding the model in selecting the relevant tables for a user's query. 
Column descriptions, on the other hand, detail the data contained in each column, helping the model to accurately map natural language to specific columns. This is particularly important when synonyms or varied phrasings are used, or when multiple fields are highly similar.

In practical applications, table and column descriptions are sometimes missing, while manually crafting descriptions for complex databases is often a labor-intensive and error-prone task that requires a comprehensive understanding of the database's context.
As a result, NL2SQL systems face a cold-start problem, where the absence of such metadata hinders the accurate generation of SQL queries. 
Therefore, automatically generating these descriptions offers an effective alternative to bridge the gap between natural language and database schema when explicit descriptions are unavailable.


To address the cold-start problem in NL2SQL tasks, we propose an automated method for generating descriptions of tables and columns by utilizing a dual-process strategy: a coarse-to-fine approach followed by a fine-to-coarse mechanism.
The coarse-to-fine approach initiates the understanding process at the broader database level, gradually narrowing down to tables and eventually individual columns.
It provides a comprehensive contextual framework, enabling a holistic understanding of the database structure before focusing on its individual elements.
On the other hand, the fine-to-coarse approach starts at fine-grained column level and then ascends through the hierarchy to tables.
It offers detailed micro-level insights, ultimately contributing to a more accurate and nuanced understanding when stepping back to consider tables as a whole. 
By integrating these two complementary approaches, we aim to achieve a more robust and effective understanding of database structures, improving the performance of NL2SQL.

To demonstrate the effectiveness of the proposed method, we conduct experiments on Bird~\cite{bird} development benchmark.
The experimental results indicate that using the generated descriptions by our method improves SQL generation accuracy by 0.98\% compared to not using descriptions. 
This accounts for bridging 37\% of the gap that is usually made by manual annotations.

\section{Database Description Generation}
\begin{figure} 
    \centering
    \includegraphics[width=0.98\textwidth]{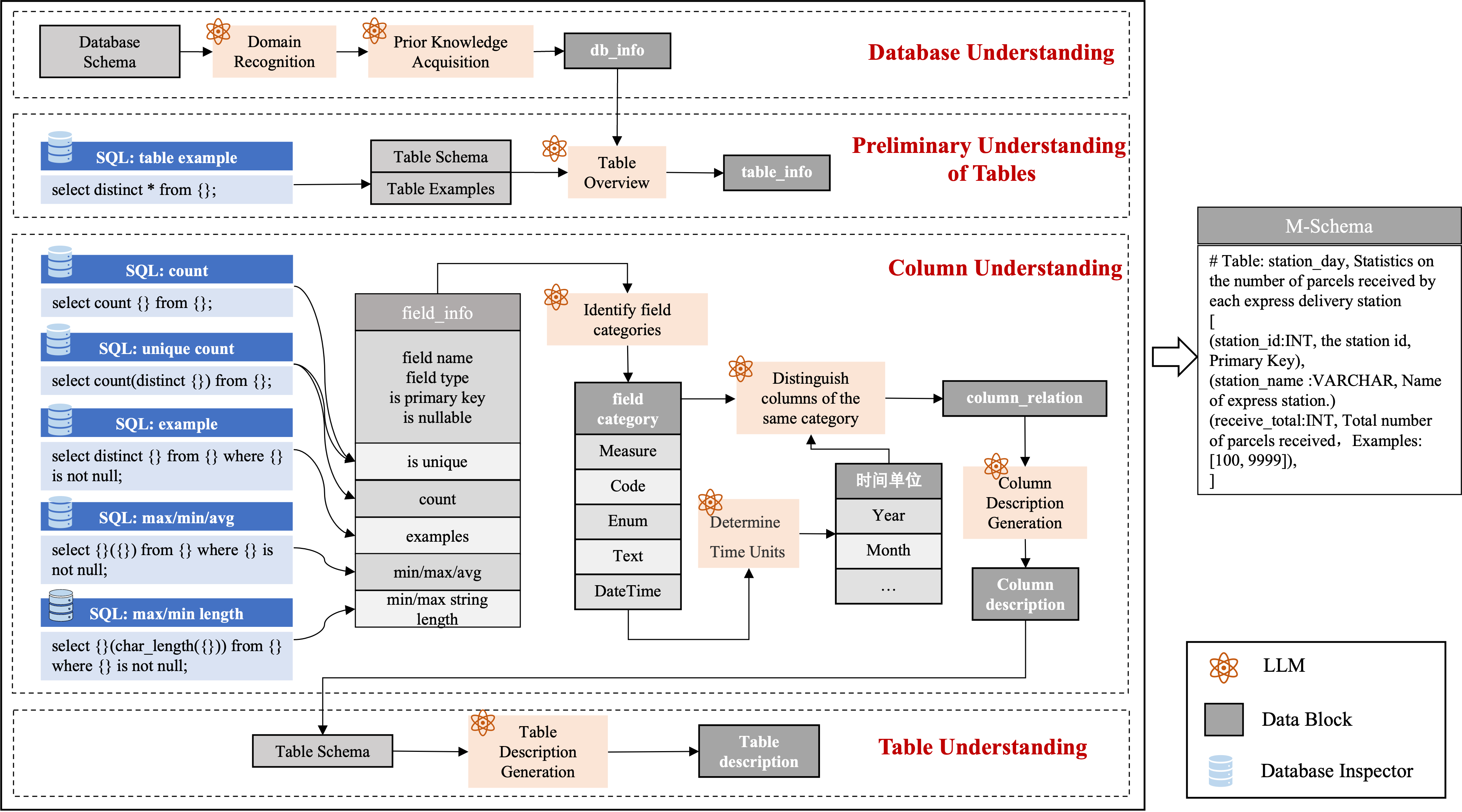}
    \caption{The workflow of proposed database description generation method.}
    \label{fig3}
\end{figure}


\subsection{Database Understanding}
The proposed methodology for database description generation begins with an overall understanding of the database. 
By providing the complete database schema to an LLM, the following objectives are achieved:

1. Domain Recognition: Identify the specific domain to which the database pertains and provide an overview of its general content. 
This step involves discerning the thematic and functional context of the database, enabling a clearer perspective for further analysis.

2. Prior Knowledge Acquisition: Utilize the knowledge acquired by the LLM during its pre-training phase to analyze typical dimensions and metrics of interest within the identified domain. 
The focus is on revealing key attributes that are relevant and insightful, guiding further investigation and leading to a more targeted understanding of the database's potential applications.

Understanding the overall architecture and interrelationships within a database provides a foundational comprehension that enables accurate and relevant insights when examining specific tables and columns. 
At the same time, it ensures that the detailed analysis remains contextually aligned. 
We represent the information obtained in this stage as db\_info, which serves as the basis for a detailed analysis of the tables and columns.

\subsection{Preliminary Understanding of Tables}

This step concentrates on analyzing each individual table within the database based on the comprehension of the database established in the previous step.
This involves functional analysis of tables and semantic prediction of columns.
First, analyze what data is stored in this table and what the function of the table might be.
Then, leverage the contextual information from $db\_info$ to hypothesize the semantic significance and potential meaning behind the data. 






By synthesizing this foundational understanding of tables denoted as $table\_info$, a detailed and coherent portrayal of each table's content, purpose, and significance is established. 
This aligns with the broader domain objectives identified earlier, and facilitates targeted analysis and insights in subsequent stages of column analysis.


\subsection{Column Understanding}

\subsubsection{Identify the category of the field}
In databases, fields are typically categorized into two categories: dimensions and measures. Dimensions are attributes that describe the data and provide a way to segment, filter, or categorize it, while measures are numerical values that can be aggregated or analyzed mathematically. To gain a deeper understanding of each field, we divide dimensions into four categories: Code, Enum, DateTime, and Text. 

\begin{itemize}
    \item Code: This category refers to fields that contain identifiers or codes, which are often used to uniquely identify an entity or object in the database, such as a user IDc.
    \item Enum: Enumerated fields are dimensions that contain a predefined list of possible values. These category typically represents categorical data, such as status (e.g., 'active', 'inactive'), or any limited and fixed set of options that describe an attribute.
    \item DateTime: Datetime fields capture data related to a specific point in time. These dimensions allow data to be organized according to different granularities, such as hours, days, months, or years. This category of fields is essential for time-related analysis, such as measuring trends and timestamps of events.
    \item Text: This category contains unstructured or semi-structured text data. These fields can be used to perform qualitative analysis based on text information such as name and description.
\end{itemize}

For each column, $column\_info$ is extracted from the database, which includes: 1) basic details: field name, field type, whether it is a primary key, whether it is nullable, is unique, and field examples; 2) statistical information: count(distinct), maximum/minimum and average values, as well as the maximum and minimum string lengths.
We present $filed\_info$ to LLM and use a series of processes to categorize the columns.
Due to the importance of DateTime dimensions in BI analysis, we also use LLM to determine the granularity of DateTime columns, especially those stored as int or string types.
This additional information is then added into $column\_info$ for 


\subsubsection{Distinguish columns of the same category}
In light of the observation that columns within the same category frequently exhibit semantic associations or similarities, for each category of columns,  
we present the entire table alongside fields of the same category to LLM, prompting it to analyze the differences and interconnections between these fields. 
The output of LLM in this phase is denoted as $column\_relation$.
This process is crucial, as it helps the model in effectively differentiating between similar fields, thereby avoiding potential confusion.


\subsubsection{Column description generation}
In this step, we use LLM to predict the likely semantics of a field based on the table's basic information $table\_info$, relationships between fields of the same category $column\_relation$, and the field's basic information $field\_info$. 
The database-level, table-level context, and inter-column relationships help the model accurately understand each column's role within the table. We've found that overly long column descriptions can hinder the effectiveness of NL2SQL, so we limit the length less than 20 words.

\subsection{Table description generation}
Obtaining the semantics of each column, we update M-Schema.
At this stage, we prompt LLM to overview the entire table, summarize its content, and analyze its potential application as a table description.
By integrating detailed insights from the column level upwards, it enables the discovery of patterns or anomalies that might otherwise be overlooked, thus refining the overall comprehension of the database.
The length of table description is limited within 100 words.

\subsection{Implementation}

We have released the source code of our proposed method on GitHub at~\url{https://github.com/XGenerationLab/XiYan-DBDescGen}. 
Our tool supports three widely used database engines: SQLite, MySQL, and PostgreSQL, and it is also compatible with M-Schema.

We provide four modes of description generation:
\begin{itemize}
    \item No\_comment: Remove all table and column descriptions.
    \item Origin: Uses the original comments directly extracted from the database's Data Definition Language (DDL).
    \item Generation: Remove all table and column descriptions and then generates entirely using a Large Language Model (LLM).
    \item Merge: Combines the "Origin" and "Generation" modes. The model generates descriptions for tables and columns that lack original comments while retaining the original comments where they exist.
\end{itemize}
You can directly connect to the database, generate table descriptions and build M-Schema, which can be used for downstream NL2SQL task.

\section{Experiments}

We conduct experiments on Bird development~\cite{bird} set to validate the effectiveness of proposed  description generation method.
The evidence is not utilized in our experiments, since evidence provides information that aids in understanding tables and fields.
To demonstrate the generalization ability of proposed description generation method, we use three widely used open source models as SQL generator,
Qwen2.5-Coder-14B~\cite{qwen25coder}, Codestral-22B~\footnote{https://huggingface.co/mistralai/Codestral-22B-v0.1} and 
Llama3.1 8B~\footnote{https://huggingface.co/meta-llama/Llama-3.1-8B-Instruct}.
The database structure is constructed as M-Schema~\cite{xiyansql} and provided to LLM.
We use Execution Accuracy (EX) to evaluate the effectiveness of the generated SQL queries.

To evaluate the impact of descriptions on NL2SQL performance, we compare three scenarios: without any descriptions, using descriptions generated by proposed method, and using manually crafted descriptions.
The results are shown in Table~\ref{table:compare}. 
On average, our generated descriptions enhance performance by 0.93\% compared to having no description and achieves 39\% of the manual level. 
This illustrates the positive effect of our description generation method on improving the performance of NL2SQL.

\begin{figure}[t] 
    \centering
    \includegraphics[width=0.9\textwidth]{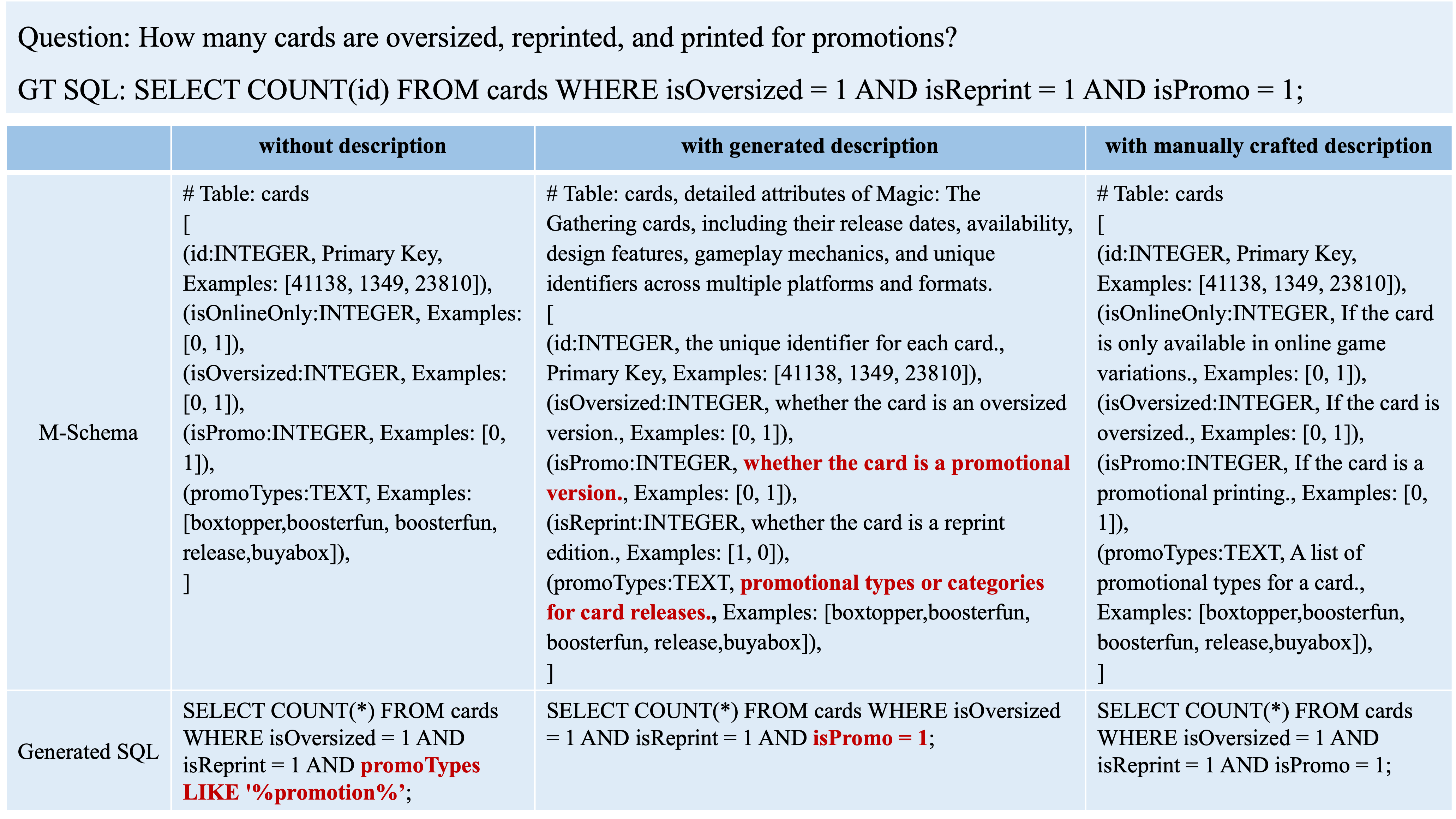}
    \caption{The impact of different database descriptions on SQL generation.}
    \label{fig:description_case}
\end{figure}

Figure~\ref{fig:description_case} illustrates how a database description influences SQL generation. 
Without the description, an LLM might misunderstand the semantic of the column "isPromo" and incorrectly link the term "promotions" with the column "promoType." 
Our proposed method, however, accurately infers that "isPromo" signifies "is promotional." 
It does this by analyzing the entire table, making connections with other fields, and integrating with the inherent knowledge of LLM, ultimately achieving correct schema linking.

\begin{table}[!tp]
 \caption{Performance comparison on Bird benchmark with different descriptions.}
    \begin{center}
    \begin{minipage}{0.95\textwidth}
    \begin{tabular}{cccc}
    \toprule
    Method & Qwen2.5-Coder-14B  & Codestral-22B & Llama3.1 8B \\    
    \midrule
    EX(w/o description, \%) & 37.82 & 39.07 & 31.87 \\
    EX(w generated description, \%) & 39.37&  39.49 & 32.68 \\
    EX(w manually crafted description, \%) & 40.18& 41.07 &  34.51 \\
    \bottomrule
    \end{tabular}
    \end{minipage}
    \end{center}
\label{table:compare}
\end{table}

\section{Conclusion}
In this report, we propose an automated method for table and column description that employs a dual-process approach: coarse-to-fine followed by fine-to-coarse. 
By integrating these two complementary approaches, we aim to achieve a more robust and effective understanding of database structures, facilitating improved translation of natural language queries into SQL.
The experimental results indicate that using the proposed method to generate descriptions improves SQL generation accuracy by 0.98\% compared to not using descriptions.

\bibliographystyle{plain}
\bibliography{references}

\end{document}